\documentclass[letterpaper]{article} 
\usepackage[submission]{aaai2026}  
\usepackage{times}  
\usepackage{helvet}  
\usepackage{courier}  
\usepackage[hyphens]{url}  
\usepackage{graphicx} 
\usepackage{natbib}  
\usepackage{caption} 
\usepackage{algorithm}
\usepackage{algorithmic}
\usepackage{array}
\usepackage{booktabs} 
\usepackage{xcolor} 
\usepackage{colortbl}
\usepackage{multirow}
\usepackage{amsfonts}
\usepackage{amsmath}

\usepackage{newfloat}
\usepackage{listings}

\DeclareCaptionStyle{ruled}{labelfont=normalfont,labelsep=colon,strut=off} 
\floatstyle{ruled}
\newfloat{listing}{tb}{lst}{}
\floatname{listing}{Listing}
\title{TAlignDiff: Automatic Tooth Alignment assisted by Diffusion-based Transformation Learning}

\author {
    Yunbi Liu\textsuperscript{\rm 1},
    Enqi Tang\textsuperscript{\rm 1},
    Shiyu Li\textsuperscript{\rm 1},
    Lei Ma\textsuperscript{\rm 3},
    Juncheng Li\textsuperscript{\rm 4},
    Shu Lou\textsuperscript{\rm 2}\\
    Yongchu Pan\textsuperscript{\rm 2},
    Qingshan Liu\textsuperscript{\rm 1}
}
\affiliations {
    \textsuperscript{\rm 1}School of Computer Science, Nanjing University of Posts and Telecommunications, Nanjing, 210003, China\\
    \textsuperscript{\rm 2}Department of Orthodontics, Affiliated Hospital of Stomatology, Nanjing Medical University, Nanjing, 210093, China\\
    \textsuperscript{\rm 3}School of Electronics and Information Engineering, Tongji University, Shanghai, 201804, China\\
    \textsuperscript{\rm 4}School of Communication and Information Engineering, Shanghai University, 200444, China
}

\usepackage{bibentry}

\begin{document}

\maketitle

\begin{abstract}
Orthodontic treatment hinges on tooth alignment, which significantly affects occlusal function, facial aesthetics, and patients' quality of life. Current deep learning approaches predominantly concentrate on predicting transformation matrices through imposing point-to-point geometric constraints for tooth alignment. Nevertheless, these matrices are likely associated with the anatomical structure of the human oral cavity and possess particular distribution characteristics that the deterministic point-to-point geometric constraints in prior work fail to capture. To address this, we introduce a new automatic tooth alignment method named TAlignDiff, which is supported by diffusion-based transformation learning. TAlignDiff comprises two main components: a primary point cloud-based regression network (PRN) and a diffusion-based transformation matrix denoising module (DTMD). Geometry-constrained losses supervise PRN learning for point cloud-level alignment. DTMD, as an auxiliary module, learns the latent distribution of transformation matrices from clinical data. We integrate point cloud-based transformation regression and diffusion-based transformation modeling into a unified framework, allowing bidirectional feedback between geometric constraints and diffusion refinement. Extensive ablation and comparative experiments demonstrate the effectiveness and superiority of our method, highlighting its potential in orthodontic treatment.
\end{abstract}

\section{Introduction}
\begin{figure}[t]
\centering
\includegraphics[width=0.45\textwidth]{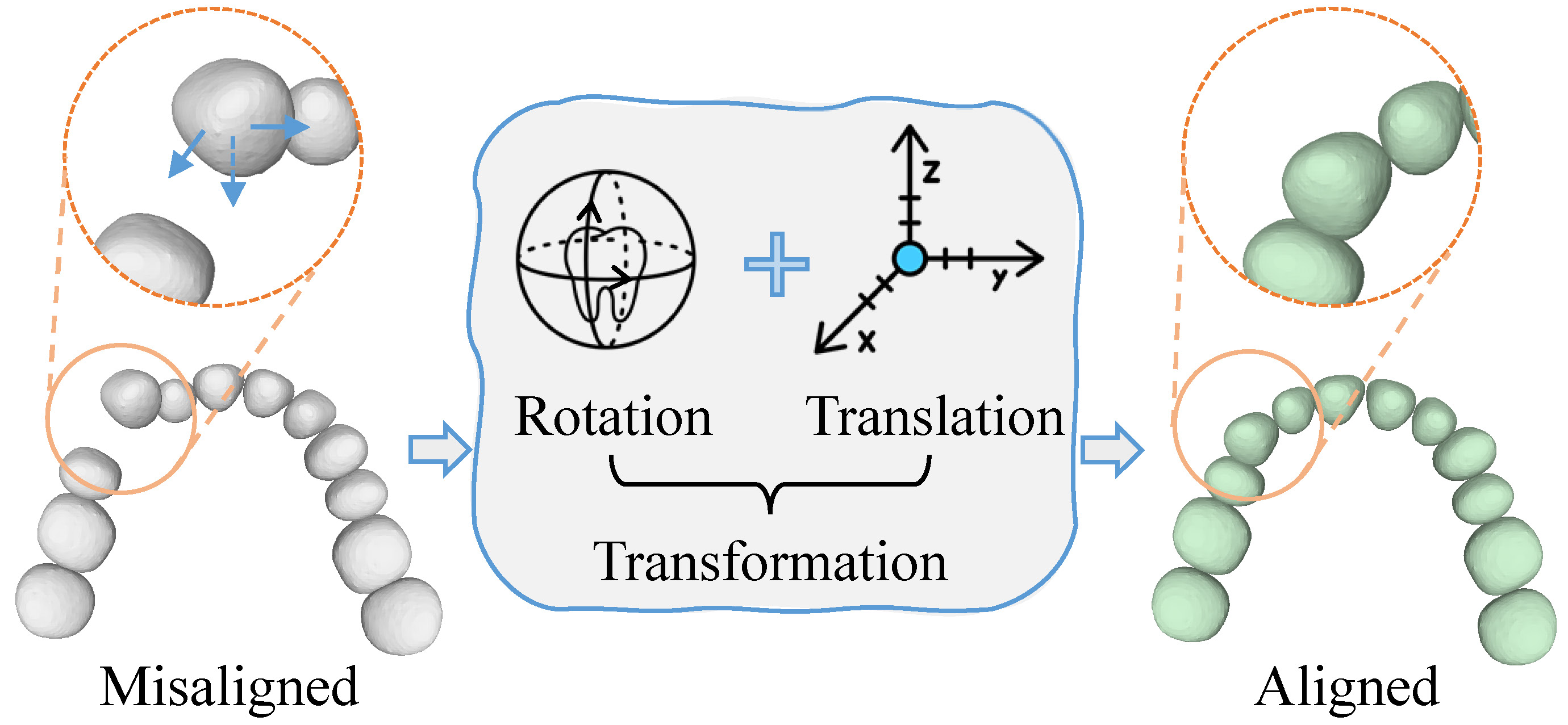}
\caption{Illustration of tooth-wise alignment, showing a grey misaligned tooth transformed into an orderly green one through rotation and translation.} 
\label{fig_1}
\end{figure}

Achieving precise tooth alignment is a fundamental objective in orthodontic care, with profound implications for occlusal function, facial aesthetics, and patient quality of life. 
However, traditional orthodontic planning and computer-aided design systems remain heavily reliant on clinicians’ experience, leading to time-consuming workflows and subjective clinical judgments~\cite{dawson2006functional,grauer2011accuracy,cheng2015personalized,kazimierczak2024ai,lei2024automatic,wang20243d}.  
Automatic tooth alignment emerges as a promising solution: by predicting optimal tooth movements, it provides clear visualizations of final outcomes, aiding clinicians in developing accurate treatment plans (Fig.~\ref{fig_1}). This approach enhances the efficiency and precision of orthodontic care, which is of great significance in clinical orthodontics.

Recent deep learning methods have advanced oral data analysis by leveraging diverse modalities, including oral photographs~\cite{lingchen2020iorthopredictor,chen2022orthoaligner,wang20243d}, dental mesh models~\cite{lei2024automatic}, and point cloud data~\cite{li2020pstn,wei2020tanet,lingchen2020italignet,lei2025teethgenerator,fan2022tad,deng2024taposenet,lei2024automatic,lei2025teethgenerator,xu2024teethdreamer,cui2019toothnet,cui2022fully}. 
Among these, tooth point cloud data provides detailed and accurate representations of dental anatomy. Current methods utilizing such data typically encode 3D tooth models into point clouds and use regression networks trained with explicit point-wise geometric losses to output optimal transformation parameters (e.g., rotation, translation parameters)~\cite{li2020pstn,wei2020tanet,deng2024taposenet,wang20243d,lei2024automatic,lingchen2020italignet}.
These methods have achieved promising automatic tooth alignment performance, significantly alleviating the burdens of traditional manual alignment.

However, geometric constraints alone on point clouds are insufficient for predicting clinically valid tooth alignment parameters, as they fail to account for the inherent distribution characteristics of transformation parameters—including valid ranges for rotation/translation and correlations between adjacent teeth. \textit{Thus, we aim to develop a method that maintains geometric consistency while simultaneously learning the inherent distribution characteristics of transformation matrices.}

Most recently, Lei et al. introduced a diffusion probabilistic model (DPM)-based framework (TADPM)~\cite{lei2024automatic}, which leverages both point cloud and mesh data to condition the diffusion process, enabling robust learning of transformation matrix distributions from malocclusion to normal occlusion. This pioneering approach inspired our research, particularly in its integration of DPMs to model the inherent variability of orthodontic data.
However, TADPM directly conditions the diffusion model on high-dimensional geometric features extracted from raw point clouds and meshes, which increases computational complexity and demands large datasets for the diffusion process. 

To address these challenges, we propose TAlignDiff, an automatic tooth alignment framework that unifies geometry-constrained transformation regression with diffusion-assisted unsupervised distribution learning. Unlike TADPM method that directly regresses transformation matrices from geometric features, our approach first predicts initial transformation matrices using a point cloud regression model (PRN). A lightweight diffusion-based model is employed to model the latent distribution of transformation matrices by progressively corrupting and denoising them. By comparing the estimated noise from the predicted and ground truth matrices, the diffusion-based model refines the regression output—with closer alignment between estimated noises indicating higher proximity to the true transformation distribution.

This design reduces the input dimensionality for the diffusion process (focusing on transformation matrices rather than raw geometric data), significantly enhancing adaptability to small datasets. By combining explicit geometric constraints from point cloud regression with implicit distribution modeling via diffusion, our method aims to provide a viable solution for clinical scenarios.

The main contributions of this work are as follows:
\begin{itemize}
    \item we propose a novel automatic tooth alignment method that integrates geometry-constrained transformation regression with diffusion-assisted unsupervised distribution learning. This design leverages both explicit geometric constraints and implicit distribution modeling.
    \item we establish bidirectional feedback between geometric regression and diffusion-based refinement. By comparing noise estimates derived from predicted and ground truth matrices, our model iteratively aligns predicted transformations with the true distribution.
    \item Extensive experiments, including ablation studies and comparisons with state-of-the-art methods, demonstrate the effectiveness and superiority of our proposed method. 
\end{itemize} 

\section{Related Work}
\subsection{Learning-based Tooth Alignment Studies}
Recent related work has demonstrated the potential of deep learning to address the complexities of tooth alignment~\cite{fan2022tad, chen2022orthoaligner,li2020pstn,wei2020tanet,deng2024taposenet,wang20243d,lei2024automatic,lingchen2020italignet}. For instance, the TANet method~\cite{wei2020tanet} innovatively employs a graph-based feature propagation module to update features extracted by PointNet, effectively tackling the six degrees of freedom (6-DOF) pose prediction problem for each tooth. 
The TAPoseNet framework~\cite{deng2024taposenet} represents a significant leap forward by integrating a multi-scale Graph Convolutional Network (GCN) to characterize teeth relationships at different levels (global, local, intersection). 
These learning-based automatic tooth alignment methods have demonstrated superior performance compared to traditional machine learning-based approaches. They offer a more accurate and efficient way to determine the optimal arrangement of teeth, which is crucial for successful orthodontic treatment planning. However, most of these methods are based on sole reliance on geometric constraints for point cloud reconstruction, overlooking the intrinsic properties of the transformation matrices. In our work, we propose a unified framework that integrates geometry-constrained transformation regression with diffusion assistance, providing a more comprehensive and accurate representation for tooth alignment.

\subsection{Diffusion Probabilistic Models}
Recent advances in diffusion probabilistic models (DPMs) have demonstrated remarkable capabilities in learning complex data distributions across diverse domains, including image synthesis~\cite{ho2020denoising,liu2024ladiffgan,wang2024structure}, 3D shape generation~\cite{luo2021diffusion,guo2025neuro}, and conditional generation~\cite{zhang2024dosediff,zhu2025cycle,wang20253d}. These models operate by gradually transforming a simple distribution, typically a Gaussian noise, into the target distribution through a process known as diffusion, which involves iteratively refining the data towards the desired output. In orthodontics, Lei et al. were the first to introduce a diffusion-based framework that learns the distribution of clinically valid tooth transformation matrices from malocclusion to normal occlusion~\cite{lei2024automatic}. By conditioning on geometric features extracted from dental meshes and point clouds, TADPM generates transformation parameters that align with biomechanical constraints, thereby improving generalization to real-world patient data. Inspired by their work, we believe that applying DPMs to automatic tooth alignment can be regarded as a promising solution. Unlike their work that directly regress transformation matrices from high-dimensional geometric features, we reduce the input dimensionality of the diffusion process by focusing on transformation matrices rather than raw point cloud or mesh data. This streamlining significantly mitigates reliance on large datasets, enhancing adaptability to small clinical datasets with limited annotations.

\begin{figure*}[t]
\centering
\includegraphics[width=0.85\textwidth]{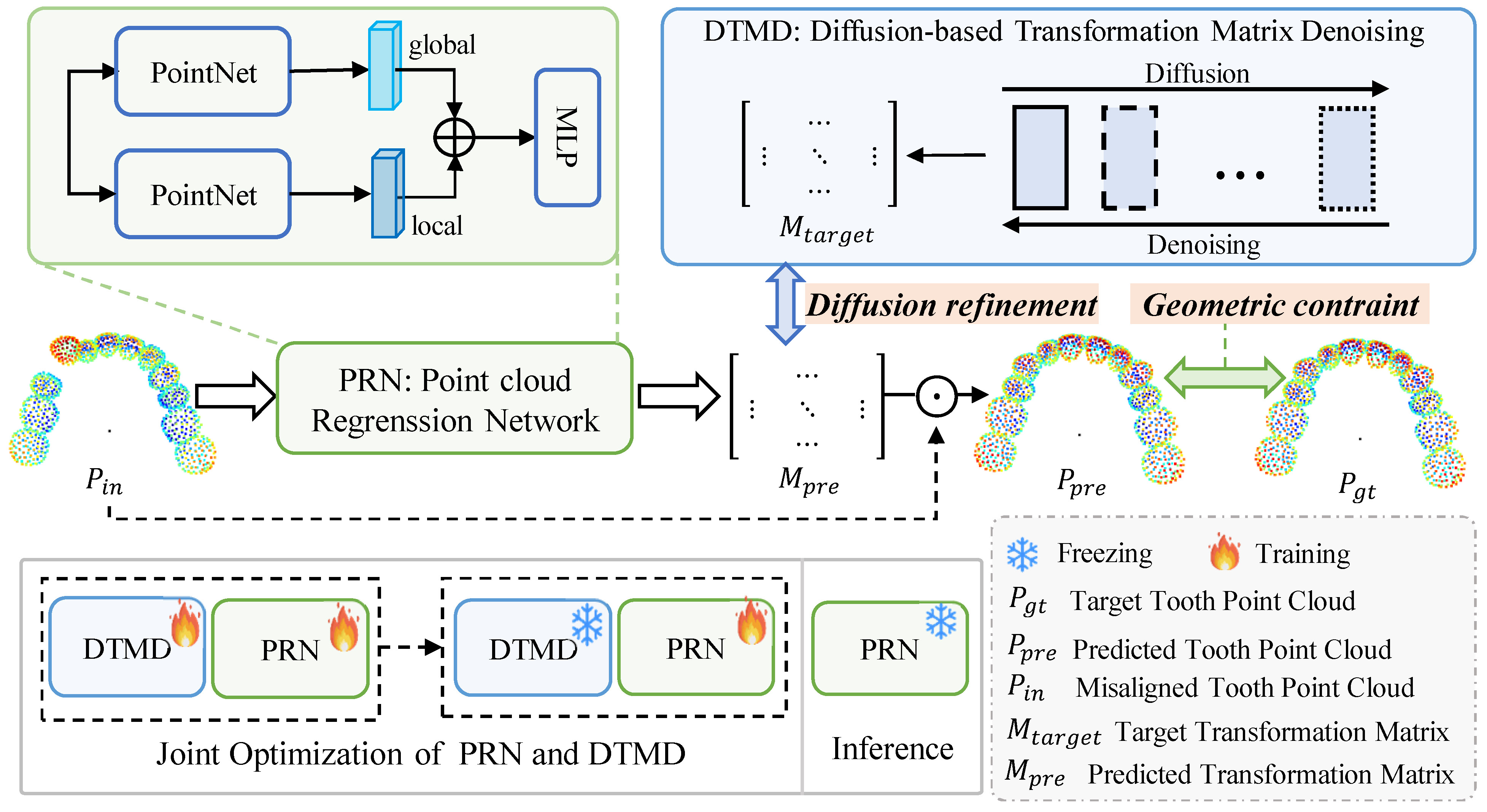}
\caption{Overview of the proposed TAlignDiff method for automatic tooth alignment. } 
\label{fig_framework}
\end{figure*}

\section{Methods}
\subsection{Problem Formulation} In this work, we aim to extract tooth-related information from 3D tooth point cloud data and positional movement details from transformation matrices to predict post-orthodontic tooth arrangement based on pre-orthodontic tooth point clouds. Let $P = \{ p_i \, | \, i = 1,...,N \} \subseteq \mathbb{R}^{N \times 3}$ be the domain of tooth point cloud data, where $N$ indicates the number of points in a $(x, y ,z)$ Cartesian space sub-sampled from dental models. Denote $P_{in}$ and $P_{gt}$ as the pre- and post-aligned tooth point cloud data, respectively. Each tooth point cloud data can be organized into $M=32$ sub-point clouds, where $M$ indicates the number of permanent teeth.
In orthodontics, the position transformation matrix $T= \{ T_i \, | \, i = 1,...,M \} \subseteq \mathbb{R}^{M \times 4 \times 4}$ serves as a crucial tool for describing the movement of each tooth. By constructing this transformation matrix, we can simplify the complex three-dimensional dental point clouds into a more intuitive mathematical form, facilitating a clearer understanding and adjustment of each tooth's position throughout the treatment.
For each tooth, its transformation matrix $T_i \in \mathbb{R}^{4 \times 4}$  can be formulated as
\begin{equation}
T_i = \begin{bmatrix}  
R_i & D_i \\
0 & 1  
\end{bmatrix} =   
\begin{bmatrix}  
R_{i11} & R_{i12} & R_{i13} & D_{ix} \\
R_{i21} & R_{i22} & R_{i23} & D_{iy} \\
R_{i31} & R_{i32} & R_{i33} & D_{iz} \\
0 & 0 & 0 & 1  
\end{bmatrix}  
\end{equation}
where $T_i$ is a  $4 \times 4$ matrix, including a $3 \times 3$ rotation matrix $R_i$ and a $3 \times 1$ displacement matrix $D_i$. The transformation matrix represents the positional movement of teeth from their initial pre-orthodontic state to their final post-treatment arrangement. For the whole unaligned tooth point cloud $P_{in}$, its corresponding aligned point cloud $P_{target}$ can be transformed by applying $T$ to $P_{in}$, \textit{i.e.,}, $P_{gt} = T \cdot P_{in}$. 
In this work, we propose a novel tooth alignment method, comprising two main modules (as shown in Fig.~\ref{fig_framework}): 1) a point cloud-based regression network (PRN) and 2) a DDPM-based Transformation Matrix Denoising model (DTMD).

\subsection{Point Cloud-based Regression Network}
The input to the Point Cloud-based Regression Network (PRN) is unaligned 3D tooth point cloud $P_{in}$, and the output is the transformation matrix $T$. We can predict the aligned point clouds $P_{predict}$ by applying the transformation matrices to the unaligned dental point clouds.
PointNet specializes in raw point clouds to extract hierarchical features efficiently using symmetric functions~\cite{qi2017pointnet}. We adopt the encoder part of PointNet as the feature extractor of tooth point clouds, which consists of three 1d convolutional layers with channels of $[64, 128, 1024]$. Two PointNet encoders, $\epsilon_g$ and $\epsilon_l$, are used to separately extract global features for the whole dentition and local features for tooth-level geometric details. The global and local features are concatenated to form the overall feature representation of the entire 3D tooth point cloud data. With the feature representation as input, an MLP-based decoder, consisting of $3$ fully-connected layers with channels of $[512, 256, 16]$, is used to regress the transformation matrices. The operation of the tooth alignment network can be formulated as:
\begin{equation}
    T^* = \phi(\epsilon_g(P_{in}) \oplus \epsilon_l(P_{in}))
\end{equation}
where $\phi(\cdot)$ and $T^*$ are the MLP-based decoder and predicted transformation matrix, respectively. $\oplus$ indicates a concatenation operation for the global and local tooth features.

\paragraph{Geometry-constrained Loss Function} We employ a point-wise reconstruction loss and a tooth centroid offset loss to ensure the geometry alignment of the predicted point cloud data and the ground truth one. By applying the transformation matrix to an unaligned tooth point cloud, we can convert it to a transformed tooth point cloud. The reconstruction loss is used to penalize the position difference between the target tooth point cloud and the predicted aligned one, which can be formulated as:
\begin{equation}
    L_{rec} = \frac{1}{N} \sum \left\|T^* \cdot P_{in} - T \cdot P_{in}\right\|_1
\end{equation}
The position changes of tooth centroids after orthodontic treatment can reflect the overall collective tooth displacement of the tooth arrangement. To effectively constrain the tooth centroid position differences between the predicted tooth point cloud and the target one, we propose the incorporation of a tooth centroid offset loss. The tooth centroid offset loss, denoted as $L_{center}$, is formulated to minimize the $L_1$ norm of the difference between the centroids of the predicted and target tooth point clouds. Mathematically, it is expressed as:
\begin{equation}
    L_{center} = \frac{1}{M} \sum \left\| C_{predict} - C_{target}\right\|_1
\end{equation}
where $M$ represents the total number of teeth under consideration, $C_{predict}$ and $C_{target}$ denote the centroids of the predicted and target tooth point cloud data, respectively. This loss serves as a significant constraint in our model, ensuring that the predicted tooth arrangement closely aligns with the desired one in terms of the centroid positions of the teeth. 

By employing a point-wise reconstruction loss and a tooth centroid offset loss, the PRN can effectively ensure the geometric alignment in post-aligned tooth point clouds, thereby regressing an accurate transformation matrix. However, the standalone PRN is insufficient to capture the inherent characteristics of the transformation matrix. To address this limitation, the proposed method introduces a diffusion-based denoising model to enhance the performance of the PRN.

\subsection{Diffusion-based Transformation Denoising Model} 
Transformation matrices encapsulate the sequential arrangement of teeth, as well as the relative angles and distances between adjacent teeth. The diffusion process utilized in our denoising model is particularly advantageous for modeling the probabilistic nature of transformation matrices, as it systematically captures and reconstructs the underlying distribution of the effective transformations. Consequently, we introduce a diffusion-based transformation matrix denoising (DTMD) model to optimize the PRN by modeling the latent distribution of transformation matrices.

The diffusion process of the transformation matrices includes the forward chain of adding noise and the reverse chain of denoising. We iteratively add Gaussian noise to $M_0$, the reshaped target transformation matrices $M_{gt}$, through the forward diffusion chain with an adaptive length $T$. Formally, we define the distributions of the transformation matrices in timestep $t$ by:
\begin{equation}
    q(M_t \mid M_0) = \mathcal{N}(M_t \mid \sqrt{\gamma_t}M_0, (1-\gamma_t) I)
\end{equation}
where $\gamma_t \in (0, 1)$ are the variances of the Gaussian noise in $T$ iterations.
The training objective of the diffusion model is defined as:
\begin{equation}
    L_{diffusion} = \mathbb{E}_{M_t, \epsilon \sim \mathcal{N}(0,1), t} \left[ \left\| \epsilon - \epsilon_{\theta_d}(M_t, t) \right\|_2^2 \right]
\end{equation}
 The term $\epsilon_{\theta_d}$ denotes the noise estimator parameterized by the diffusion model $\theta_d$, which predicts the noise embedded in the noisy transformation matrices of the target domain.

\paragraph{Diffusion-based Loss Function} We propose a novel contrastive denoising loss to enhance the stability of the PRN. By comparing noise estimates derived from predicted and ground truth matrices, our model aligns predicted transformations with the true distribution.
The contrastive denoising loss, denoted as $L_{denoi}$, is defined as:
\begin{align}
L_{denoi} =  \mathbb{E}_{M_{gt}^t, M_{pre}^t, t} \left[ \left\|\epsilon_{\theta_d}(M_{gt}^t, t) - \epsilon_{\theta_d}(M_{pre}^t, t) \right\|_1  \right] 
\end{align}
where $M_{gt}^t$ and $M_{pre}^t$ denote the ground truth and predicted transformation matrices at time step $t$ after Gaussian noise has been added, respectively. The $L_1$-norm measures the discrepancy between the noise estimates, encouraging alignment between the noise distributions of $M_{gt}^t$ and $M_{pre}^t$. By minimizing the noise discrepancy, the diffusion model indirectly corrects the predicted transformation matrix, guiding it toward the target distribution.

\subsection{Joint Optimization Framework}
As mentioned above, the proposed method integrates two key components: the point cloud-based transformation regression network (PRN) and the di
The PTA model utilizes geometry-constrained losses to provide supervision within the point cloud, enabling it to effectively represent and reconstruct the spatial configuration of teeth. 
Simultaneously, the DTMD model functions to capture the complex distributions of the transformation matrices. By estimating the noise characteristics of these matrices, the DTMD model enhances the reliability of the predictions made by the PRN model. This synergy allows the diffusion model to refine the predictions of the transformation matrix, ensuring that they are aligned with realistic biological and mechanical constraints of orthodontic tooth alignment. In general, the proposed method leverages both explicit geometric constraints and implicit distribution modeling. The overall optimization objective is defined as follows:
\begin{equation}
    L_{total} = L_{rec} + \lambda_1 L_{center} + \lambda_2 L_{denoi} + \lambda_3 L_{diffusion}
    \label{eq7}
\end{equation}
where $\lambda_1$, $\lambda_2$, and $\lambda_3$ are weighting factors that balance the contributions of these loss functions during training.
This joint optimization framework through the hybrid loss function and a staged training strategy (as shown in Fig.~\ref{fig_framework}), promotes an intricate interplay between the geometric features of the tooth point cloud and the latent distribution of the transformation matrices, enabling more effective and reliable orthodontic predictions.

\subsection{Training Details}
Data augmentation and a joint training strategy are key training steps in implementing our method. 
\paragraph{Data Augmentation} To expand the diversity of training samples, we introduce a data augmentation approach, which is crucial given the relatively small number of samples available for training. The data augmentation method involves multi-tooth rotation and single-tooth translation. 
\begin{itemize}
    \item Multi-tooth Rotation. We randomly select $k$ teeth ($5 \leq k \leq 10$) from the complete dentition and apply independent Euler-angle rotations about their local coordinate axes for the $i$-th selected tooth, adhering to biomechanical constraints of dental movement. This simulates natural variations in tooth orientation that might occur due to individual differences in dental anatomy.
    \item Single-tooth Translation. Individual teeth are randomly translated within the three-dimensional space of the point cloud. This mimics the displacement of teeth that could result from orthodontic treatment.
\end{itemize}
Note that to ensure label consistency with augmented inputs, we compute the inverse transformation matrices for all manipulated teeth.  
\paragraph{Joint Training} 
We adopt a staged joint training strategy for the PRN and DTMD models in the proposed method.
As shown in Fig.~\ref{fig_framework}, initially, both the PRN and DTMD are jointly trained for the first $200$ epochs to facilitate collaborative learning and ensure both models achieve stability. Subsequently, in the following $200$ epochs, only the PRN model is trained, leveraging the pre-trained DTMD to optimize the PRN model's output, which reduces computational resources and enhances the transformation matrix prediction capabilities of the PRN model, all while maintaining the DTMD parameters fixed based on the pre-training results. Our method is implemented with PyTorch and trained on a server equipped with one $3,090$ GPU. We use Adam optimizer with a learning rate of $0.01$ for the PRN model and a slower one of $0.005$ for the DTMD model. The batch size and the number of training epoch are set to $4$ and $400$, respectively. 
Note that the DTMD only participates in the training process and does not contribute to the inference stage of the model. Thus, although DTMD provides crucial optimization supervision, its computational overhead does not affect inference efficiency. This lightweight design ensures that the model maintains high effectiveness during the inference phase while enhancing performance during training through collaborative optimization. For clarity, our code will be publicly available once accepted.

\begin{figure}[t]
\centering
\includegraphics[width=0.45\textwidth]{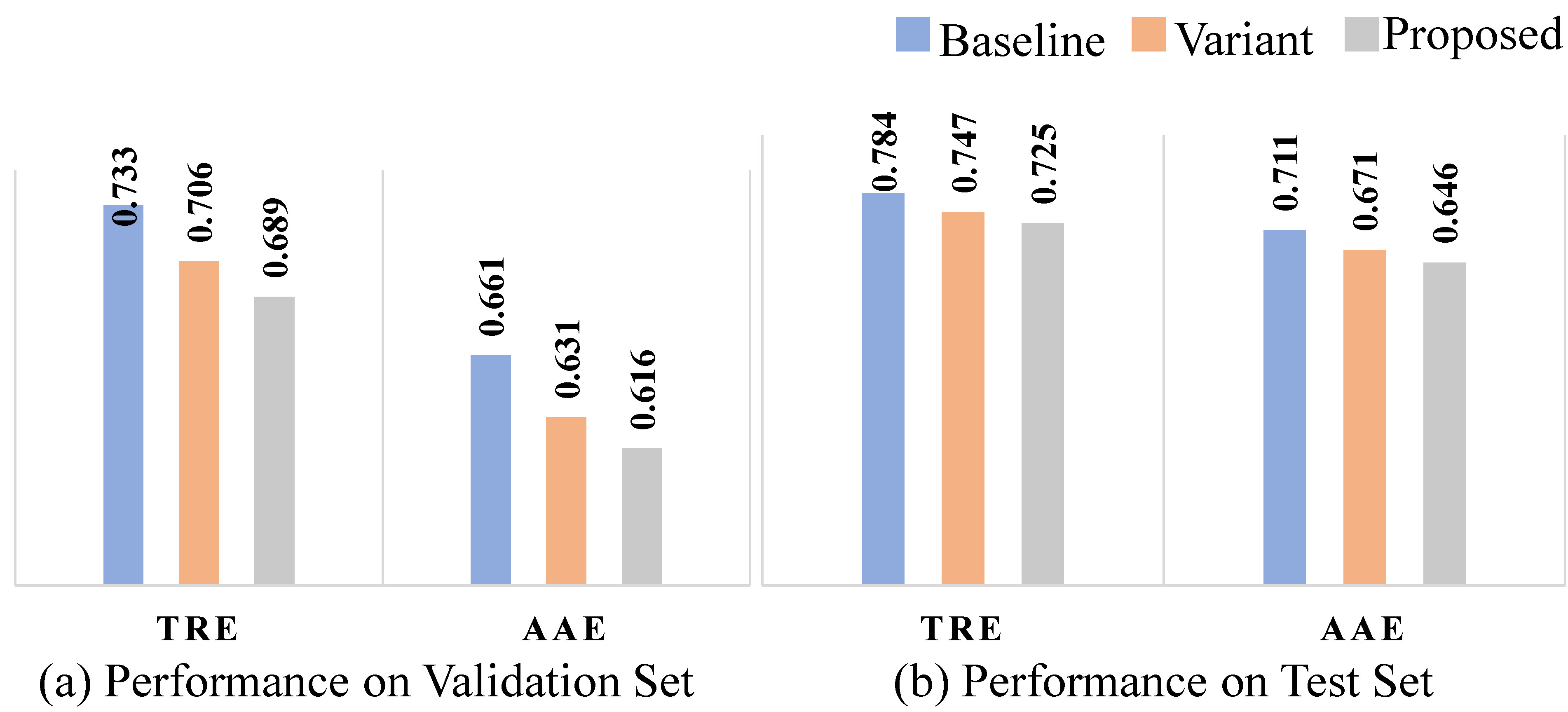}
\caption{Performance Histograms of Baseline, Variant, and the Proposed Method. } 
\label{fig_histogram}
\end{figure}
\begin{figure}[t]
\centering
\includegraphics[width=0.45\textwidth]{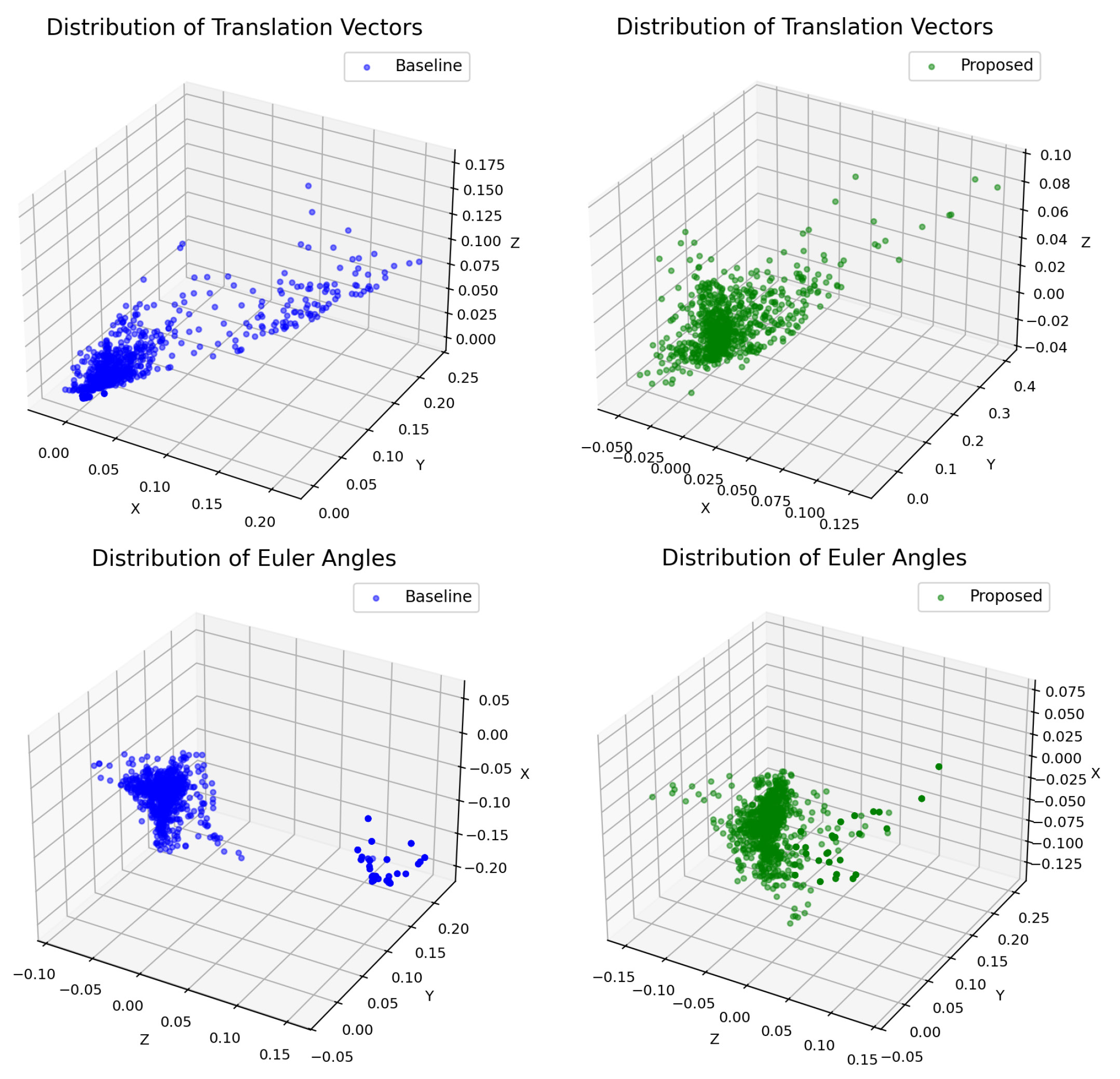}
\caption{3D scatter plots illustrating the distribution of predicted transformation matrices (translation vectors and Euler angles) for the baseline model (blue) and the proposed method (green) on the test set.} 
\label{fig_distribution}
\end{figure}

\section{Experiments and Results}
\subsection{Dataset Description} 
Our study utilized a dataset sourced from the Automatic Tooth Arrangement Challenge at the 7th International Symposium on Image Computing and Digital Medicine (ISICDM 2024)~\footnote{http://www.imagecomputing.org/isicdm2024/}. This dataset encompasses pre-orthodontic treatment tooth point cloud data from $124$ patients, gathered in a clinical environment. Each tooth's point cloud consists of $4,096$ points, which are structured into 32 sub-point clouds. Notably, each sub-point cloud comprises $128$ points, randomly sampled from oral scans of the 32 permanent teeth.
In addition to the point cloud data, the dataset also includes treatment plans formulated by orthodontists. These plans are encoded as 3D transformation matrices, with each matrix being a $4\times4$ configuration, leading to an overall shape of $32\times4\times4$ for each input sample. As per the challenge guidelines, the dataset was split into training, validation, and test sets, adhering to a ratio of 74:20:30.

\begin{table}[t]
\centering
\resizebox{0.48\textwidth}{!}{
\begin{tabular}{cccccc}
\toprule
\multicolumn{3}{c}{Loss Weight}
& \multirow{2}{*}{Dataset} & \multirow{2}{*}{TRE} & \multirow{2}{*}{AAE} \\
\cline{1-3}
$\lambda_1$ &$\lambda_2$ &$\lambda_3$ & & & \\ \hline
0 & 0 & 0 & Validation & $0.733 \pm 0.826$ & $0.661 \pm 0.740$ \\
& & &Test & $0.784 \pm 0.927$ & $0.711 \pm 0.831$ \\ \hline
0.1 &0 &0 & Validation & $0.712 \pm 0.795$ & $0.638 \pm 0.715$ \\
& &  &Test & $0.748 \pm 0.873$ & $0.670 \pm 0.778$ \\ \hline
0.1 &0.01 &0.1 & Validation & $\textbf{0.690} \pm \textbf{0.751}$ & $\textbf{0.617} \pm \textbf{0.662}$ \\
&  &  &Test & $\textbf{0.725} \pm \textbf{0.834}$ & $\textbf{0.646} \pm \textbf{0.734}$ \\ \hline
0.2 &0.01 &0.1 & Validation & $0.734 \pm 0.811$ & $0.695 \pm 0.733$ \\
& &  &Test & $0.766 \pm 0.864$ & $0.692 \pm 0.766$ \\ \hline
0.1 &0.05 &0.1 & Validation & $0.710 \pm 0.782$ & $0.638 \pm 0.696$ \\
&  &  &Test & $0.739 \pm 0.850$ & $0.663 \pm 0.755$ \\ \hline
0.1 &0.005 &0.1 & Validation & $0.694 \pm 0.774$ & $0.623 \pm 0.686$ \\
& & & Test & $0.725 \pm 0.834$ & $0.646 \pm 0.734$ \\ 
\bottomrule
\end{tabular}}
\caption{Performance Comparison with Different Loss Weights.}
\label{tab1}
\end{table}

\subsection{Experimental Settings}
We conducted multiple experiments to evaluate our method's effectiveness. First, an ablation study is performed to validate the effectiveness of key factors, the DTMD module and joint optimization strategy, used in the proposed method. The point cloud-based transformation regression with only point-wise reconstruction loss is regarded as the baseline method.
Then, we conducted a group of experiments to investigate how varying the coefficients $\lambda_1$, $\lambda_2$, and $\lambda_3$ impacts the tooth alignment results.
Additionally, we evaluate our method against two state-of-the-art tooth alignment methods, PSTN and TADPM, as well as two widely-used models tailored for point cloud data, including PointNet++ and PointMLP, to highlight its superiority.

Note that we have trained two widely used models specifically designed for point cloud data, namely PointNet++ and PointMLP, as our PRN model with geometric-constrained losses, for the sake of fairness. These models were trained with identical training objectives and parameter settings. Furthermore, we trained the tooth alignment method, TADPM, using the open-source code available on GitHub \url{https://github.com/lcshhh/TADPM}. And we reproduced the PSTN method based on the implementation details in the paper~\cite{li2020pstn}.
\paragraph{Metrics} Target registration error (TRE) is a common metric that measures the distance between predicted and target tooth point clouds. We also introduced absolute arch error (AAE), a new metric that quantifies the difference between the predicted and target post-arranged dental arches.

\subsection{Ablation Study}
To evaluate the DTMD module and joint optimization strategy, we performed an ablation study with a baseline, variant, and proposed model.  The variant model included the PRN and DTMD modules, trained separately initially. Performance comparison of these models is presented in the histograms in Fig.~\ref{fig_histogram}. The proposed model demonstrated superior results, as evidenced by the lowest TRE and AAE metrics. Compared to the baseline method, our proposed and variant models offered distinct advantages over the baseline approach due to the integration of the DTMD module. Our joint training approach further outperformed the variant, enhancing performance through collaborative optimization.

We convert the predicted transformation matrices by the baseline and the proposed methods on the test set into a 3D translation component and a 3D Euler angle component. Figure~\ref{fig_distribution} visualizes the 3D scatter plots of their respective translation and rotation components. Each point represents the translation values or Euler angles of a translation matrix for one tooth. As can be observed, the proposed method generates better-clustered translation matrices than the baselines, suggesting improved model stability.

\subsection{Impact of the Loss Weight}
Hybrid loss functions are used in our proposed method to refine the model's performance. 
The performance analysis of our proposed model, as detailed in Table~\ref{tab1}, yields several key observations regarding the impact of loss weight.
\textit{Firstly}, the incorporation of the centroid loss ( $L_{center}$ ) markedly enhances model performance, particularly when its corresponding weight ($\lambda_1$) is set to 0.1. This improvement is significant compared to the scenario where all weights are set to zero, indicating the beneficial role of $L_{center}$. \textit{Secondly}, the inclusion of the contrastive denoising loss ( $L_{denoi}$) contributes to further performance improvements. This enhancement is observed when the DTMD model is trained with both $L_{diffusion}$ with the weight of 0.1 and $L_{denoi}$ with the weight of 0.01. These findings validate the effectiveness of the proposed DTMD module in enhancing the model's predictive capabilities. \textit{Finally}, the proposed model demonstrated its best performance when the weight factors $\lambda_1$, $\lambda_2$, and $\lambda_3$ were set to 0.1, 0.01, and 0.1, respectively.
In summary, the strategic assignment of weights to different loss components within our proposed model is crucial for achieving superior performance in automatic tooth alignment tasks. The results underscore the importance of $L_{center}$ and $L_{denoi}$ in refining the model's ability to predict accurate tooth movements and transformations.

\begin{figure*}[th!]
\centering
\includegraphics[width=0.90\textwidth]{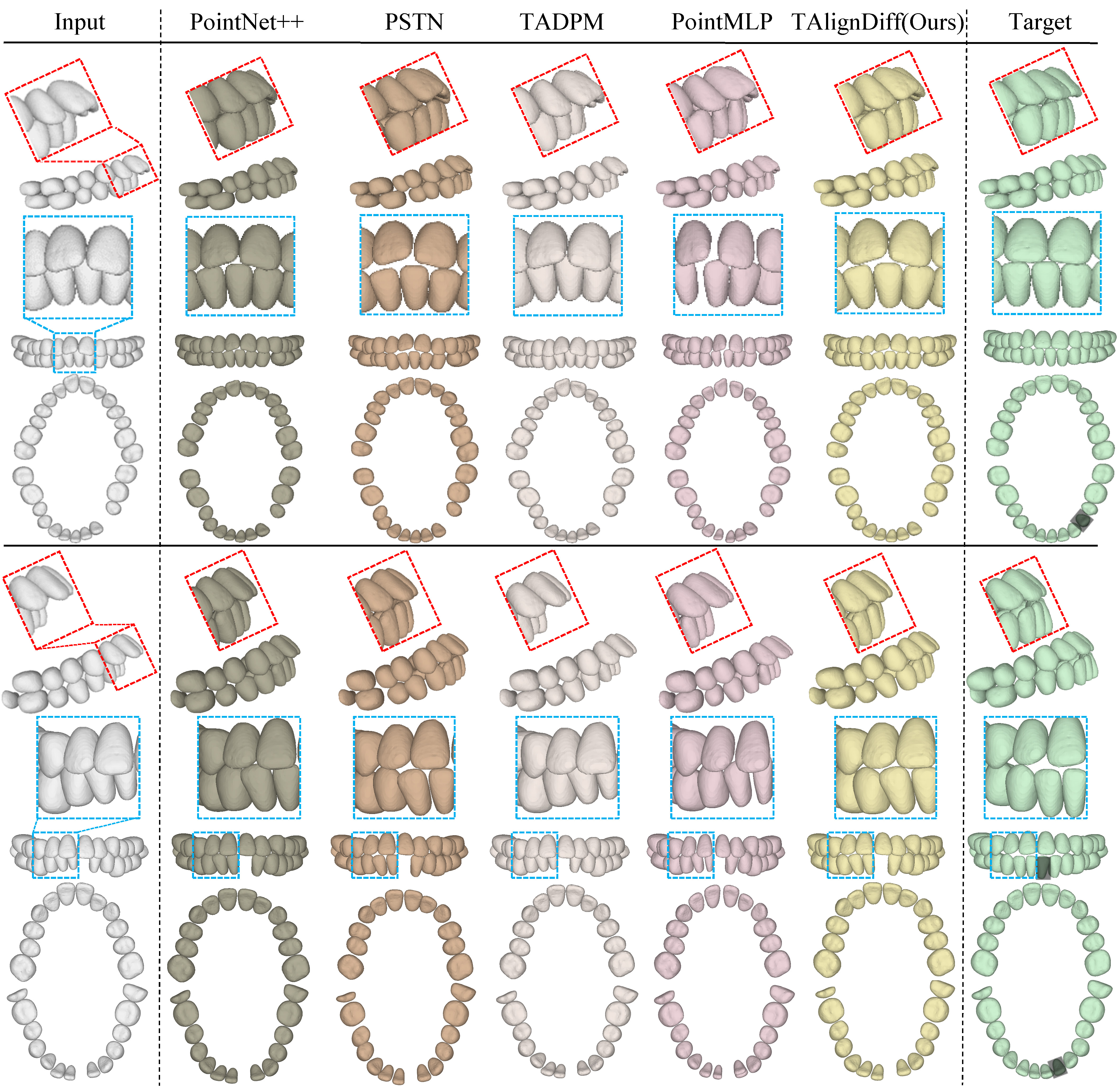}
\caption{Mesh reconstruction results of various methods. For clarity, we show the reconstructed mesh results of different views. The red and blue dashed rectangles highlight specific teeth for focused comparison. Note that the teeth within the black box under the target represent implanted teeth that are missing in the input.} 
\label{fig_visual}
\end{figure*}
\begin{table}[t]
\centering
\resizebox{0.45\textwidth}{!}{
\begin{tabular}{lccc}
\toprule
Model & Dataset & TRE & AAE \\
\hline
PointNet++ & Validation & $0.769 \pm 0.860$ & $0.702 \pm 0.782$ \\
& Test & $0.791 \pm 0.927$ & $0.717 \pm 0.833$ \\
\hline
PointMLP & Validation & $0.826 \pm 0.935$ & $0.758\pm 0.866$ \\
& Test & $0.819 \pm 0.935$ & $0.743 \pm 0.844$ \\
\hline
TADPM & Validation & $0.907 \pm 0.982$ & $0.848 \pm 0.922$ \\
& Test & $0.890 \pm 0.963$ & $0.821 \pm 0.883$ \\
\hline
PSTN & Validation set & $0.730 \pm 0.822$ & $0.658 \pm 0.736$ \\
& Test set & $0.779 \pm 0.917$ & $0.705 \pm 0.821$ \\
\hline
Proposed & Validation & $\textbf{0.690} \pm \textbf{0.751} $& $\textbf{0.617} \pm \textbf{0.662}$ \\
& Test & $\textbf{0.725} \pm \textbf{0.834}$ & $\textbf{0.646} \pm \textbf{0.734}$ \\
\bottomrule
\end{tabular}}
\caption{Performance Comparison with various Methods.}
\label{tab2}
\end{table}

\subsection{Comparison with the State-of-the-art Methods}
To substantiate the efficacy of our proposed method, a comparative analysis was conducted against several contemporary approaches. The quantitative outcomes of this analysis are detailed in Table~\ref{tab2}, which presents the performance metrics of various models on both the validation and test sets. Furthermore, we provide visual illustrations of the tooth arrangement outcomes by different methods as depicted in Fig.~\ref{fig_visual}. For clarity, we reconstruct the mesh results from the aligned tooth point cloud results predicted by various methods.
Our proposed method achieved the lowest values for target registration error (TRE) and average arch error (AAE), with a notable improvement over other methods, as evidenced in Table~\ref{tab2}(with p value less than 0.01). The visual results in Fig.~\ref{fig_visual} validates the quantitative findings, with the proposed method's aligned point clouds closely matching the target. Three deep overbite malocclusion cases are visualized, where mesh reconstructions show varying improvement across methods. The proposed method achieves superior visual alignment, closely resembling the target dental formation, especially in handling deep overbite scenarios.

In contrast, the performance of other comparison methods declined in these typical cases, likely due to poor adaptability to small samples. For instance, TADPM—an inspiration for our work—regresses transformation matrices via diffusion using high-dimensional point cloud features as input~\cite{lei2024automatic}. Its performance is thus limited by the small sample size in this work. In contrast, our method performs initial predictions in point cloud space using PointNet, coupled with auxiliary optimization via a DTMD model, thereby reducing the sample size requirement.

\section{Conclusion}
This study introduces a novel automatic tooth alignment framework, a unified framework that integrates geometry-constrained transformation regression with diffusion-based translation modeling. It consists of a point cloud-based translation regression network (PRN) and a diffusion-based translation matrix denoising module (DTMD). By combining explicit geometric constraints via PRN with implicit distribution modeling via DTMD, the proposed method yields more accurate and reliable tooth arrangement outcomes, demonstrating superior performance to existing methods on a clinical dataset. This enhancement is crucial for improving treatment planning and outcomes in orthodontics, benefiting both orthodontists and patients.
In the future, we intend to incorporate clinical context to refine our method and validate its performance on larger datasets as they become available.

\bibliography{aaai2026}

\end{document}